\documentclass[runningheads]{llncs}

\usepackage[T1]{fontenc}

\usepackage[T1]{fontenc}
\usepackage{amsmath}
\usepackage{booktabs}
\usepackage{graphicx}
\usepackage{cite}

\usepackage{tabularray}
\usepackage{indentfirst}
\usepackage{algorithm}
\usepackage{algorithmic}

\usepackage{booktabs}
\usepackage{graphicx}
\usepackage{array}
\usepackage{url,wrapfig}
\usepackage{hyperref}
\usepackage{booktabs}
\usepackage{array}
\usepackage{booktabs}

\usepackage{colortbl} % Add this package for \arrayrulecolor

\usepackage{array}
\usepackage[table,xcdraw]{xcolor} % For coloring

\usepackage{amsmath, graphicx, xcolor, colortbl} % 数学符号和颜色支持
\begin{document}

\title{An LLM Agent-Based Complex Semantic Table  Annotation  Approach}
\author{Yilin Geng\inst{1,\#}  \and
Shujing Wang\inst{1,\#}  \and
Chuan Wang\inst{1,\#} \and
Keqing He\inst{3}  \and
Yanfei Lv\inst{2,*} \and
Ying Wang\inst{1,*}\and
Zaiwen Feng\inst{1,4,5,*} \and
Xiaoying Bai\inst{2}
}
\renewcommand{\thefootnote}{}
\footnotetext{${^\#}$These authors contributed to the work equally.}
\footnotetext{${^*}$Corresponding authors.}

\authorrunning{Y. Geng et al.}
% %
% \authorrunning{F. Author et al.}
% % First names are abbreviated in the running head.
% % If there are more than two authors, 'et al.' is used.
% %
\institute{College of Informatics, Huazhong Agricultural University, Wuhan, Hubei, China, 430070 
\and
Military Science Information Research Center, Academy of Military Sciences, Beijing, 100080
\and
School of Computer, Wuhan University, Wuhan, Hubei, China, 430072
\and
Hubei Key Laboratory of Agricultural Bioinformatics
\and
Engineering Research Center of Agricultural Intelligent Technology, Ministry of Education}

\maketitle              % typeset the header of the contribution
\begin{abstract}
The Semantic Table Annotation (STA) task involving Column Type Annotation (CTA) and Cell Entity Annotation (CEA) tasks, maps table contents to ontology entities, playing important roles in various semantic applications. However, complex tables often pose challenges such as semantic loss of column names or cell values, strict ontological
hierarchy annotation, homonyms, spelling errors, abbreviations, which hinder the accuracy of annotation. To tackle these issues, this paper proposes an LLM-based agent approach for CTA and CEA tasks. We design and implement five external tools with tailored prompts based on the ReAct framework, enabling the STA agent to dynamically select suitable annotation strategies based on different table characteristics. The experiments are conducted on the Tough Tables and BiodivTab datasets related to the aforementioned challenges from the SemTab challenge, where it outperforms existing methods in various metrics. Furthermore, by using \textit{Levenshtein distance} to reduce redundant annotations, we achieve a 70\% reduction in time costs and a 60\% reduction in LLM token usage, providing an efficient, cost-effective solution for STA task.

\end{abstract}
\section{Introduction}

Tabular data, abundant in enterprise databases and the web, contains rich semantic information. Semantic Table Annotation (STA) plays a crucial role in knowledge graph construction\cite{jentab}, Ontology-Based Data Access (OBDA)\cite{SEDAR}, and data lake governance. Column Type Annotation (CTA) and Cell Entity Annotation (CEA) are the critical tasks in STA\cite{liu2023tabular}. CTA maps table columns to classes of the ontology, and CEA links cell values to specific entities. However, challenges arise in the practical task of CTA and CEA for complex tables (as shown in Figure 1), such as semantic loss of column names or cell values, ontological hierarchy strict annotation, homonyms, spelling errors and abbreviation\cite{toughtable}. Existing STA methods primarily rely on external knowledge bases and complex hardware resources, often facing challenges related to data complexity and resource consumption, particularly lacking flexibility when handling diverse tabular scenarios\cite{kgcode,lowresource}.
% Specifically, by establishing a mapping between tabular data and knowledge graphs (or ontology), STA  converts the rich information in tables into structured knowledge, which aids in knowledge graph construction and enhancement. By establishing mappings between ontological concepts and actual data sources, Ontology-Based Data Access (OBDA) resolves data integration and privacy issues across multiple data sources. Semantic annotation of tables is a key approach for semantic mappings of OBDA. Furthermore, by addressing data ambiguity caused by inconsistent terminology across systems within a data lake, semantic annotation improves data quality, enhances data interoperability, and facilitates data lake governance.

% 现有方法解决方案

% 介绍自己的课题
% In recent years, the large language models (LLMs) have demonstrated strong capabilities in semantic understanding. The LLM-based agents, such as the ReAct framework\cite{yao2023react}, use iterative reasoning and action (thought-action-observation) with external tools invocation to perform complex tasks. 
To overcome these challenges, this paper proposes an LLM agent-based semantic annotation method for CTA and CEA tasks based on ReAct framework\cite{yao2023react}. The preprocessing mechanism for error correction and abbreviation expansion, as well as five specialized tools are integrated into this STA agent to allow dynamic selection suitable annotation strategy based on the table context, enhancing semantic annotation accuracy.

Two datasets are selected from the SemTab challenge, Tough Tables and BiodivTab\cite{toughtable}\cite{biodiv}, which contain the aforementioned challenges and are annotated with DBpedia (an ontology-based public knowledge graph). Superior performance compared to existing methods was demonstrated using these datasets, achieving CTA F1-score of 0.596 and CEA F1-score of 0.843 on Tough Tables, as well as CTA F1-score of 0.89 and CEA F1-score of 0.90 on BiodivTab. Additionally, we conducted ablation experiments to evaluate the contribution of each tool and analyze their necessity in solving the semantic annotation challenges. Moreover, given the highly similar strings in the datasets, we used \textit{Levenshtein distance} to reduce redundant annotations, saving 70\% in time costs and 60\% in LLM token usage. Our method provides an automated, efficient and low-cost solution for semantic annotation.

The contributions of this study include the following three aspects:

\begin{itemize}

\item A ReAct-based agent approach is proposed to dynamically select different tool composition strategy based on the table characteristics to address CTA and CEA tasks.

\item Five tools are designed in the STA agent to tackle key challenges in CTA and CEA. Our method outperforms existing approaches across various CTA and CEA metrics on the Tough Tables and BiodivTab.

\item By utilizing \textit{Levenshtein distance} to reduce redundant annotations, the approach achieves a 70\% reduction in time costs and a 60\% reduction in LLM token usage.
\end{itemize}

\section{Challenges in the CTA and CEA}
\subsection{Problem Definition}
CTA is the process of mapping a table's columns to classes of an ontology, which aims to assign a semantic meaning to each column by linking it to the appropriate class in the ontology. For example, Figure \ref{problem_example} (a)(b)(c) links the columns to the corresponding classes of the ontology. CEA is the process of linking the cells to entities in the ontology. Figure \ref{problem_example} (d)(e)(f) link the cells to the corresponding entities of the ontology. CEA helps in recognizing the actual instances of classes in the ontology, enabling data to be queried semantically.

\begin{definition}
Column Type Annotation (CTA): Given a table \( T \) with columns $C = \{ c_1, c_2, \dots, c_n \},$ the CTA task involves predicting the semantic type(s) for each column \( c_k \in C \). This is represented as a set of ontology classes $S_k = \{ st_1, st_2, \dots, st_a \},$ where each \( st_i \) denotes a specific class in the ontology.
\end{definition}

\begin{definition}
Cell Entity Annotation (CEA): For a table \( T \) with cells $E = \{ e_{i,j} \mid 1 \leq i \leq m, 1 \leq j \leq n \}$, the CEA task aims to identify and link each cell \( e_{i,j} \) to one or more entities in the knowledge graph. This is denoted as a set of entities $E_{i,j} = \{ e_1, e_2, \dots, e_b \}$,where each \( e_i \) corresponds to a distinct entity in the ontology.
\end{definition}

\subsection{Challenges of CTA and CEA}
%在实际的表格中，因为各种原因会出现表格质量问题，所以会有以下挑战。
In CTA and CEA tasks of practical tables, the following challenges are faced, as illustrated in the Figure \ref{problem_example}:
\begin{figure}[h]
\includegraphics[width=1\textwidth]{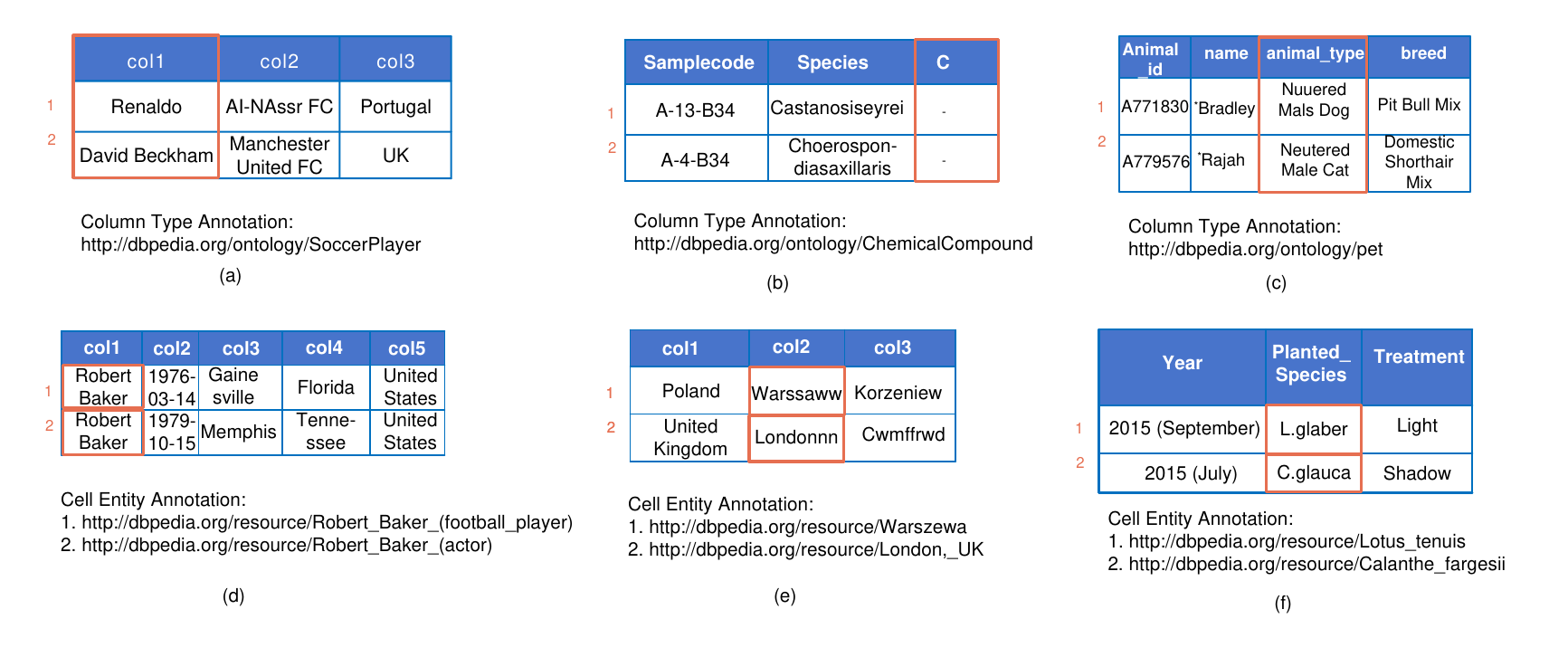}
\caption{The challenge faced by CTA and CEA.}  (a) Semantic loss of column names. (b) Semantic loss of cell values. (c) For the column "animal\_type" the strict ontology hierarchy is "pet", neither too broad such as "animal" nor too narrow such as "dog or cat". (d) The two "Robert Bakers" refer to two different individuals. (e) Spelling errors in the tables. (f) Abbreviations resulting from the domain-specific items hamper the understanding of the cell values.
\label{problem_example}
\end{figure}
\vspace{-10pt}

% 三种情形。
\textbf{Semantic loss of column names or cell values}: Semantic loss in column names or cell values hinders annotation. As is shown in Figure \ref{problem_example} (a), the column header “col1” is semantically vague, making interpretation difficult. This study uses the Column Topic Detection Tool to predict the column topic based on the cell values, replacing the original meaningless column name. As is shown in Figure \ref{problem_example} (b), the cells are null. To assist in grasping the column’s semantics, other column names of the table will be provided to the LLM as supplementary information in Context-Supported CTA Selection Tool.

\textbf{Ontological hierarchy strict annotation}: In CTA tasks, the annotation system needs to provide an appropriately hierarchical annotation based on the table's content. The semantic scope should be neither too broad nor too narrow. For example, "animal\_type" in Figure \ref{problem_example} (c) cannot simply be annotated as "animal". By examining the adjacent columns, especially the left column "name", it can be found that "pet" is more suitable, because common animals (unlike pets) are not given personal names. To solve the challenge, Knowledge Graph-Based Enhancement Tool provides enough CTA candidates, and Context-Supported CTA Selection Tool uses the adjacent columns as context information in selection, so that the LLM can choose the most appropriate column type.

\textbf{Homonyms}: There are many same-name phenomena across different fields, such as the same personal names and place names may refer to different entities in different contexts. As is shown in Figure \ref{problem_example} (d), the table contains multiple instances of "Robert Baker", referring to two different entities – one is a football player and the other is an actor. It is difficult to distinguish them based solely on the name. To solve the challenge, Context-Supported CEA Selection Tool leverages the semantics of the cell as well as other cells in the same row to distinguish same-named entities and annotate them accurately.

\textbf{Spelling errors and abbreviations}: Spelling errors and abbreviated terms are common in tables, which hamper the semantic understanding( Figure \ref{problem_example} (e)). This issue can be resolved through context-based abbreviation expansion, and data preprocessing module requires the LLM to examine the cells combining contextual information in order to correct misspelled words and complete abbreviated terms.

\section{An LLM-based Agent Approach for CTA and CEA}

\subsection{Framework of the ReAct-based Agent}

\begin{figure}[h]
\includegraphics[width=\textwidth]{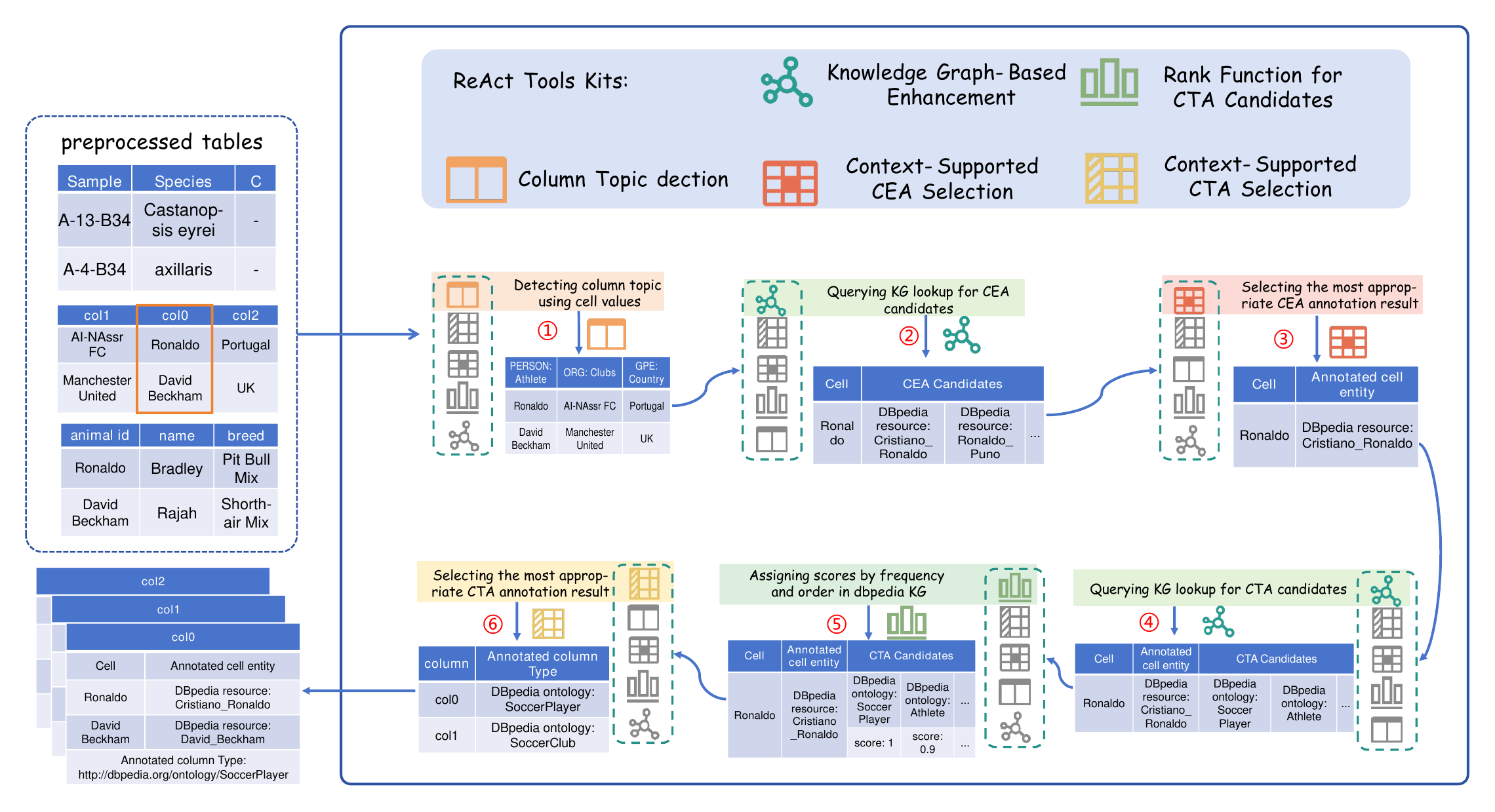}
\caption{Framework of the ReAct-based Agent for CTA and CEA.} 
\label{frame}
\end{figure}
\vspace{-5pt}
 
This study proposes a ReAct-based semantic annotation agent, with its framework illustrated in Figure \ref{frame}. This method takes a preprocessed table as input and outputs an annotated table with labels. Column names and cell values are the main content to understand the semantics of the column. Some tables have meaningful column names and cell values, which is the best situation for understanding the column. However, there is a common phenomenon of semantic loss in column names or cell values. The two situations hinder the annotation for the CTA and CEA. For the above three situations, this study trains the agent to learn three types of solutions by using prompt engineering (The prompt can be found in Supplement Figure 1 ). The ReAct-based agent dynamically selects appropriate workflows for CTA and CEA tasks based on column and cell characteristics, ensuring broad applicability without dataset-dependent modifications.
%列名，单元格值是理解该列语义的主要内容，但实际表格数据中可能会存在语义缺失的情况，通过对语义缺失的分类，本研究将表格数据分为三种类型，分别是列名语义缺失、单元格语义缺失和列名和单元格无语义缺失。本研究使用少样本学习方法，训练智能体面对三种类型的解决方案。在面对实际的表格数据时，根据不同的语义缺失情况，调用不同的工作流。
In the STA task, we adopt the ReAct framework for dynamic task decomposition and multi-tool collaboration. The ReAct framework uniquely integrates the planning capabilities of LLMs with external tool execution, effectively addressing complex table annotation challenges.

% Compared to the Chain-of-Thought (CoT) method, which relies solely on internal model knowledge and suffers from potential error propagation, ReAct enhances annotation reliability via external knowledge access and dynamic adjustment through its "planning-execution-observation" mechanism. For example, it can automatically invoke the Column Topic Detection tool when encountering the semantic loss of column name, which is obviously superior to static frameworks in handling heterogeneous data.

In terms of efficiency and robustness, the ReAct framework optimizes resource utilization and reduces the high computational overhead brought caused by the exhaustive method, through result caching and the early termination strategy based on confidence. Its iterative feature supports error detection and recovery, enabling it to maintain high system stability when dealing with noisy data. Overall, the ReAct framework has significant advantages in knowledge reliability, process flexibility, complex problem-solving ability, and system robustness. 
%代码调用图片

\subsection{Tools in the LLM-Based Agent for CTA and CEA}
\vspace{-5pt}
\subsubsection{Data Preprocessing}

% In this research, we focus on processing CSV files for accurate DBpedia queries. Spelling errors and abbreviations are corrected using LLMs to ensure text standardization. In the CTA task, duplicate removal and selecting the top K cells for annotation improve data quality. Named Entity Recognition (NER) with SpaCy identifies entities in columns, skipping those with specific types like amounts or dates to enhance downstream tasks like topic prediction and entity linking. NER also identifies domain-specific entities, improving preprocessing efficiency and column semantics for DBpedia queries. Data 
%在表格数据中，拼写错误和缩写是常见的问题，特别是专业名词的缩写，对准确理解单元格语义造成了极大的障碍，本研究通过大模型进行数据预处理解决这个挑战。通过设计提示模板，要求大模型检查单元格是否有拼写错误或者缩写，并结合单元格上下文信息进行纠正和缩写补全。数据预处理通过提升数据完整性和标准度，为正确语义标注提供基础。本研究通过命名实体识别辅助筛选拼写错误和缩写的单元格。SpaCy工具能够识别该列单元格的实体类型，本方法根据实体频度决定该列的实体类型，大模型以该列实体类型为参考，依次判断该列单元格，并筛选出不符合的单元格作为可疑的单元格，下一步对这些可疑单元格详细判断是否有拼写错误或者是否需要缩写补全。另外，为了能够通过最少的文本长度准确理解列语义，在获取该列实体类型和列主题检测中，本研究在通过去重挑选该列的代表性单元格，通过这种方法消除列前面单元格内容一致进而导致数据冗余的情况，不但能够节省token，并且代表性单元格能够最大程度地反映出该列的语义。数据预处理是提升数据质量的重要步骤，能够有效解决标注注释中拼写错误和缩写带来的挑战。
Spelling errors and abbreviations are common problems in tabular data, which severely hinder understanding of cell semantics. This study addresses this challenge through data preprocessing using the LLMs. We designed prompt templates which instruct the LLM to examine cells for spelling errors or abbreviations, and perform corrections and abbreviation expansions based on the cell context. By enhancing data completeness and standardization, this data preprocessing provides a foundation for accurate semantic annotation.

In addition, Named Entity Recognition (NER) is employed to assist in filtering cells containing spelling errors and abbreviations. SpaCy is employed to identify the entity types of cells in a column. We determine the predominant entity type for a column based on entity frequency. The LLM then employs the entity type as a reference to select cells that are inconsistent with this type for further processing. 

Furthermore, this study implements a process of deduplication to select representative cells for NER of columns and the Column Topic Detection Tool, which eliminates data redundancy arising from consistent content in the initial rows of a column, thereby not only reducing the usage of LLMs tokens but also ensuring that the representative cells effectively reflect the column semantics.

\vspace{-10pt}
\subsubsection{Column Topic Detection}
This tool addresses the semantic loss of column names by using LLMs to analyze cell data and infer meaningful column topics. Specifically, the LLM-based agent classifies tables into predefined types, and when column names lack semantics but cells contain sufficient meaningful data, it automatically invokes Column Topic Detection to replace ambiguous column names with inferred topics. The prompt can be found in Supplement Figure 2.

%列名语义缺失是表格数据的常见情况，对理解列语义产生极大的障碍，是表格数据语义标注的一个挑战。为了解决这个挑战，本研究使用大模型根据该列单元格内容检测该列主题，并将生成的列主题代替原来无意义的列名。具体来说，对于预处理后的待标注表格，大模型根据表格结构提取表格特点，从三种表格类型中选择一类，对于列名语义缺失但单元格足够且有意义的情况，智能体自动调用列主题检测工具，根据单元格内容生成列主题，并代替原来无意义的列名。提示模板如图。

\vspace{-10pt}

\subsubsection{Knowledge Graph-Based Enhancement}
% 耿
% dbpedia外部查询
% DBpedia是一个开源的，存储结构化数据的知识图谱，它将维基百科中的数据结构化地提取出来，涵盖了诸如人物、事件、组织、书籍、地理等多个领域。并且它为这些数据创建一个组织管理方式，它顶层本体，用以模式级和实例级的数据管理。DBpedia的本体包括大量Classes和Properties，包括Person, Place, Organisation等。本文DBpedia API 封装为一个工具，ReAct通过调用该工具可以进行基于本体的模式级别和实例级别的查询，为CTA和CEA任务生成候选集。网址：https://lookup.dbpedia.org/
DBpedia is an open-source knowledge graph tool that constructs knowledge graphs by extracting structured data from Wikipedia. In our research, the input of the tool is textual content of cells, and the output is ontology or resource URL, encompassing entities, categories, properties. The role of DBpedia is to provide extensive background knowledge for CTA and CEA tasks, mitigating hallucinations in LLMs, thereby improving the accuracy of entity alignment and classification. In the study, the DBpedia API is encapsulated into a tool, which is invoked by ReAct to perform ontology-based schema-level and instance-level queries, generating candidate sets for the CTA and CEA tasks.

\vspace{-10pt}
\subsubsection{Rank Function for CTA Candidates}
The tool is used to score and rank the CTA candidates. The DBpedia Lookup tool is invoked to generate \textit{K} candidate sets using the first 10 cells of the column, with each candidate set containing 10 candidate classes. The tool then scores and ranks each candidate entity by considering both its frequency and order of occurrence. The formal description is as follows:

$C(c_{1},c_{2},...,c_{10})$ is the top 10 cells set in the column. For every $c_{i}$, the DBpedia Lookup tool is used to  query the ontology classes. The top 10 classes are selected as the $candidate(i) = (can_{i1},can_{i2},...,can_{i10})$, corresponding score = $(1,0.9,...,0.1)$. For $can_{ij}$, 

\begin{equation}CTAscore(can_{ij}) = \sum_{i=1}^{n} score_i(can_{ij})
\end{equation}
 
\subsubsection{Context-Supported CEA Selection}
% 该工具调用大模型对从一个单元格的候选集中选出最终注释结果.将该单元格、该单元格所在行的其他单元格、该单元格所在列的列名以及候选集交给大模型，提示模板如下:

The tool invokes the LLMs to select the final annotation result from the candidate set of a cell. The cell, other cells in the same row, the column name of the cell, and the candidate set are provided to the LLMs. The prompt can be found in Supplement Figure 3.
During the annotation process, it is common to find duplicate cells in tabular data, where some are annotated and others are not. However, the system still goes through the annotation process for these unannotated cells, which significantly increases the time cost. The table cell annotation algorithm based on the \textit{Levenshtein distance} is designed to solve the problem. Since the annotation results involve string similarity, the Levenshtein distance is particularly suitable for handling cases where strings are similar or semantically equivalent, making it the core calculation method. Algorithm\ref{alg:levenshteindistance} calculates the \textit{Levenshtein distance} between unannotated and annotated cells to determine whether existing annotations can be reused. For cells where no suitable reusable annotation is found, a predefined annotation strategy is applied. The distance\_threshold is calculated as \textit{k} times the minimum string length between the annotated cell and the unannotated cell. Based on extensive experimentation, the optimal value for \textit{k} was determined to be 0.2. The configuration ensures an effective balance between precision and recall in the annotation process. By using the algorithm, all cells in the table can be annotated efficiently and accurately, improving the speed at which the model processes tabular data and greatly reducing the time overhead.

\textit{Levenshtein distance} is a metric for measuring the similarity between two strings, representing the minimum number of edit operations required to transform one string into another. If the lengths of two strings \(a\) and \(b\) are denoted by \(|a|\) and \(|b|\) respectively, then their \textit{Levenshtein distance} is \(\text{lev}_{a,b}(|a|, |b|)\), which satisfies:

\begin{equation}
\text{lev}_{a,b}(i,j) = 
\begin{cases} 
\max(i,j) & \text{if } \min(i,j) = 0, \\
\min
\begin{cases}
\text{lev}_{a,b}(i-1,j)+1 \\
\text{lev}_{a,b}(i,j-1)+1 \\
\text{lev}_{a,b}(i-1,j-1)+1_{(a_i \neq b_j)}
\end{cases} & \text{otherwise.}
\end{cases}
\end{equation}

Here, \(1_{(a_i \neq b_j)}\) is an indicator function that equals 0 when \(a_i = b_j\) and 1 otherwise. \(\text{lev}_{a,b}(i,j)\) represents the \textit{Levenshtein distance} between the first \(i\) characters of \(a\) and the first \(j\) characters of \(b\). (The indices \(i\) and \(j\) start from 1.)

\begin{algorithm}[!htb]
    \renewcommand{\algorithmicrequire}{\textbf{Input:}}
    \renewcommand{\algorithmicensure}{\textbf{Output:}}
    \caption{Table Cell Annotation Based on \textit{Levenshtein Distance}}
    \label{alg:levenshteindistance}
    \begin{algorithmic}[1]
        \REQUIRE 
            Table $T$, a table where some cells are annotated.
        \ENSURE 
            Table $T_{annotated\_cell}$, a table with all cells annotated.
        \STATE $T_{annotated\_cell} \gets T$
        \FORALL{$cell$ in $T_{annotated\_cell}$}
            \IF{$cell$ is annotated}
                \STATE \textbf{continue}
            \ENDIF
            \STATE $found\_similar \gets \text{False}$
            \FORALL{annotated\_cell  in $T_{annotated\_cell}$}
                \STATE $distance\_threshold \gets \min(\text{length}(cell), \text{length}(annotated\_cell)) \times 0.2$
                \STATE $d \gets \text{levenshtein\_distance}(cell,annotated\_cell)$
                \IF{$d < distance\_threshold$}
                    \STATE $T_{annotated\_cell}[cell] \gets$ annotation of $annotated\_cell$
                    \STATE $found\_similar \gets \text{True}$
                    \STATE \textbf{break}
                \ENDIF
            \ENDFOR
            \IF{not $found\_similar$}
                \STATE Annotate $cell$ using a predefined annotation strategy
                \STATE $T_{annotated\_cell}[cell] \gets annotation$ % 将新标注添加到目标表中
            \ENDIF
        \ENDFOR
        \RETURN $T_{annotated\_cell}$
    \end{algorithmic}
\end{algorithm}

\subsubsection{Context-Supported CTA Selection}
% 该工具调用大模型对从一列的候选集中选出最终注释结果.根据得分排序选取前K个本体类作为CTA决策的筛选候选集，要求大模型根据该列的单元格从筛选候选集中选取最终的本体类作为CTA的注释结果。提示模板如下:

The tool invokes the LLMs to select the final annotation result from the candidate set of a column. Based on the score ranking, the \textit{top-K} ontology classes are selected as the filtered candidate set for the CTA decision. The LLM is then required to choose the final ontology class from the filtered candidate set based on the column's cell data as the CTA annotation result. The prompt can be found in Supplement Figure 4.
\vspace{-10pt}

\subsection{Annotation Workflows for Three Types of Tables}

For the situation (Figure \ref{problem_example} (a)), column name lacks semantics, but the column contains sufficient and valid cells, which is the most common and challenging scenario. The workflow of the situation is shown in Figure \ref{frame}. The Column Topic Detection Tool  (\textcircled{1}) are firstly selected to generate a column topic based on the cell data, replacing the original meaningless column name, which is a crucial method to address the challenge of "Semantic loss of column names".  As is shown in Figure \ref{frame}, the column name is "col0", but the cells are "Ronaldo" and "David Beckham". So the cells are provided for the Column Topic Detection Tool to generate the meaningful column topic "Athelete", which is used to replace the original column name. Next, the Knowledge Graph Lookup Tool (\textcircled{2}) is selected to query the matched entity of each cell in the DBpedia knowledge graph. Because multiple entities may be matched for one cell, the \textit{top-K} matching entities will be selected in the candidate set for the cell. In Figure \ref{frame}, the matched entities of "Renaldo" are "Cristiano\_Ronaldo", "Ronaldo\_Puno" and so on, so the  \textit{top-K} of them are chosen in the candidate set. After that, the cell values, other cells in the same row, the column name, and the candidates are provided to the LLMs, which are tasked with selecting the most appropriate entity from the candidates to complete the CEA task  (\textcircled{3}). The supplement information is helpful to precisely grasp the semantics of the column and distingish the synonyms, which is an important way to address the challenge of  "ontological hierarchy strict annotation" and “synonym”.  For example, the cell "Renaldo", the other cells in the row ("AI-Nassr FC", "Portugal"),  and the column topic ("Athelete") are provided for LLMs to choose the most suitable entity ("Cristiano\_Ronaldo") from the candidates. For the CTA task, the \textit{top-K}  cells in the column are used to generate the candidates, where experiments have shown that selecting the first 10 cells is enough to understand the column's content. The annotated entity of a cell is used to query the \textit{top-M}  (\textit{M}=5/10/15) ontology classes in the DBpedia knowledge graph as a part of the candidate set (\textcircled{4}). The ontology classes derived from all \textit{K}  cells are combined to form the full candidate set. For the col0 in Figure \ref{frame}, the annotated cell entities of the first cell is "Cristiano\_Ronaldo," which belongs to ontology class "SoccerPlayer", "Athelete" and so on. These ontology classes of \textit{top-K}  cells consist of CTA candidates. A scoring function is applied to rank these candidate classes, which takes into account both the frequency and sequence of ontology classes (\textcircled{5}). The CTA score is computed for each candidate ontology class, and the \textit{top-K} ontology classes are selected based on their scores. The details of scoring are illustrated in  Section 3. Finally, the LLM selects the final ontology class from the candidate set based on the cell values (\textcircled{6}).  For the col0 in Figure \ref{frame}, the annotated column type of the column and the annotated entities are stored in the annotation result table.
\vspace{-1pt}

For the situation shown in Figure \ref{problem_example} (b), where the column name has semantics but the cell values of the column are meaningless. It doesn't need to implement the CEA task. However, for CTA tasks, the loss of cell values renders the External Query Tool unusable and the lack of CTA candidates, because the column names don't adequately represent the column semantics, especially in the situation that the column name is an abbreviation, such as the query results of column name "C" in the DBpedia graph mostly relating to "ProgrammingLanguage". For this situation, this method finishes CTA task in step \textcircled{6}. This method incorporates the column name along with other column names in the table as supplementary information, requiring the LLMs to give an ontology class as the annotated CTA label, which resolves the problem of limited contextual information caused by the loss of cell values.

For the situation shown in Figure \ref{problem_example} (c), where the column name has semantics and the column contains sufficient and valid cell values, the method skips the step \textcircled{1} and proceeds with the subsequent workflows\textcircled{2}-\textcircled{6} to complete the CTA and CEA tasks.
% 这种情况是难度较高的，本方法重点关注这种情况，使用该类型数据集，解决该场景下的应用。
% \begin{figure}[h]
% \includegraphics[width=\textwidth]{prompt.png}
% \caption{The prompts of the ReAct and tools.} 
% \label{prompt}
% \end{figure}
% \vspace{-20pt}
\vspace{-10pt}

\section{Evaluation}
\subsection{Datasets}
To comprehensively evaluate the performance of the proposed method in our study, two representative datasets, namely Tough Tables and BiodivTab, were selected for the experimental validation.

\textbf{Tough Tables} The dataset includes 180 tables with 16,464 entities and 663,830 matches. All tables lack column names, and data spans multiple domains and languages. Key challenges include name ambiguity, spelling errors, structural complexity, and noisy tables, making it ideal for testing real-world robustness.

\textbf{BiodivTab} The dataset includes 50 tables with biological fields, containing specimen observation data, numerical dominance, and abbreviations/special formats, which increase the complexity of entity matching. In the dataset, the column names have clear meanings, while some cell values lack explicit significance. To address the issue, the model is provided with headers, initial column values, and complete header information, leveraging its semantic understanding capabilities.
\subsection{Evaluation Metrics}
We evaluate the experimental performance using two metrics: Precision and F1-score, defined as follows:

\begin{equation}P=\frac{|\text{Correct Annotations}|}{|\text{System Annotations}|}, R=\frac{|\text{Correct Annotations}|}{|\text{Target Annotations}|}, F1=\frac{2\times P\times R}{P+R}\end{equation}

\subsection{Performance of the ReAct-based approach}

\begin{table}[h]
\centering
\caption{ Comparison with baselines}
\label{tab:comparison}
\begin{tabular}{@{}cllllcccc@{}}
\toprule
&   \multicolumn{4}{c}{Tough Tables}&\multicolumn{4}{c}{BiodivTab}  \\
\cmidrule(lr){2-5} \cmidrule(lr){5-9}
&   \multicolumn{2}{c}{CTA}& \multicolumn{2}{c}{CEA
}&\multicolumn{2}{c}{CTA} &\multicolumn{2}{c}{CEA} \\
\midrule
Our System&   F1 & Pr & F1 &Pr &F1 & Pr & F1 & Pr \\
\cmidrule(lr){1-9}
Our System (Gemini)&   \textbf{0.596}& \textbf{0.629}& \textbf{0.843}&\textbf{0.845}&\textbf{0.89}& \textbf{0.89}& \textbf{0.90}&\textbf{0.93}\\
 Our System (GPT-4o-mini)& 0.583& 0.613& 0.817& 0.823& 0.87&0.88& 0.89& 0.90\\
Our System (DeepSeek)& 0.585& 0.617& 0.821& 0.827& 0.88&0.88& 0.90& 0.91\\
\cmidrule(lr){1-9}
KGCODE-Tab &   \textbf{0.480} & 0.485& \textbf{0.827}&0.830 &\textbf{0.87}& \textbf{0.87}& \textbf{0.91}& \textbf{0.91}\\
 TSOTSA&   0.342& \textbf{0.627}& 0.595&\textbf{0.957}
&0.79& 0.79& 0.76&0.76
\\
 JenTab&   
0.234& 0.290& 0.572&0.796
&0.41& 0.42& 0.55&0.61
\\
 s-elBat &   0.373 & 0.375 & 0.789 &0.808 
&0.00& 0.00& 0.06& 0.06
\\
Kepler-aSI&   0.154& 0.154& -&-&0.73& 0.78& 0.53&0.53\\
 DAGOBAH& -& -& -& -& 0.62& 0.62& -&-\\
 
\toprule
\end{tabular}
\end{table}

Table \ref{tab:comparison}  shows the execution performance of the system in our study, with different models, and other annotation systems on the Tough Tables and BiodivTab datasets, which were awarded in SemTab 2022. We evaluate performance using F1-score and Precision for CTA and CEA tasks (best results in bold). Gemini slightly outperforms other models due to its extensive context window (up to 1 million tokens). Meanwhile, the system in our study performs excellently in the CTA task of the Tough Tables dataset, with the best F1 - score reaching 0.596 and the best precision being 0.629, which is significantly better than other annotation systems. In the CEA task of the Tough Tables dataset, the precision is slightly lower than that of the TSOTSA system. The lower F1-score, coupled with the improved precision of the TSOTSA system, might reflect a trade-off where the system avoids annotating difficult table cells, thus potentially increasing precision. Besides, when compared with other systems, both the F1 value and the precision are the best. Besides,in the CTA and CEA tasks of the BiodivTab dataset, the various indicators of the system are also ahead of other systems, demonstrating good performance advantages.

\subsection{Ablation Study}
% 我们进行消融实验探究每个工具对方法效果的影响。实验结果显示，知识图谱查询对系统性能影响很大、候选集数量选取K=10时，系统效果最好，添加列主题检测对系统性能有些许提升。 单独对于CTA和CEA来说，在数据预处理时对列中单元格去重，能提升CTA的性能，CEA Based on Levenshtein Distance能够极大提升系统效率，减少成本。
Ablation studies were conducted to explore the impact of each tool on the system's performance. The experimental results indicate that the Knowledge Graph Lookup has a significant effect on system performance, selecting K=10 for the candidate set yields the best results, and incorporating column topic detection slightly improve the performance. For CTA and CEA individually, deduplicating cells of the columns during Data Preprocessing enhances CTA's performance, and CEA based on \textit{Levenshtein Distance} significantly improves system efficiency.

% 2、调用外部工具vs 直接查询。实验结果和实验分析
\subsubsection{Knowledge Graph Lookup}
Table \ref{tab:dbpedia_comparison} shows the execution performance of two configurations, with and without external knowledge graph. The optimal results of CTA and CEA are presented in bold. Ablation results show that integrating external KG lookup significantly improves performance compared to relying solely on LLMs.

% \vspace{-10pt}
\begin{table}[h]
\centering
\caption{Impact of Knowledge Graph Lookup on System Performance}
\label{tab:dbpedia_comparison}
\begin{tabular}{@{}ccccccccc@{}}
\toprule
&   \multicolumn{4}{c}{Tough Tables}&\multicolumn{4}{c}{BiodivTab}  \\
\cmidrule(lr){2-5} \cmidrule(lr){5-9}
&   \multicolumn{2}{c}{CTA}& \multicolumn{2}{c}{CEA
}&\multicolumn{2}{c}{CTA} &\multicolumn{2}{c}{CEA} \\
\midrule
&   F1 & Pr & F1 &Pr &F1 & Pr & F1 & Pr \\
\cmidrule(lr){1-9}
Our system with KG Lookup& \textbf{0.596}& \textbf{0.629}& \textbf{0.843}& \textbf{0.845}&\textbf{0.89}&\textbf{0.89}&\textbf{0.90}&\textbf{0.93}\\
Our system without KG Lookup&   0.275& 0.301& 0.796&0.797&0.83& 0.83& 0.82& 0.86\\

\toprule
\end{tabular}
\end{table}

\vspace{-20pt}

\subsubsection{Number of Candidates}

Table \ref{tab:k_comparison} shows F1-scores and precision for different candidate numbers (1, 5, 10, 15) in CTA and CEA tasks. Increasing candidate numbers from 1 to 10 improves performance, with optimal results at 10 candidates. More candidates introduce noise and overhead. Our system efficiently achieves high performance with fewer candidates than other methods (e.g., KGcode’s 50), benefiting from LLMs' internal knowledge.

% \vspace{-10pt}
\begin{table}[h]
\centering
\caption{Impact of candidate number on System Performance}
\label{tab:k_comparison}
\begin{tabular}{@{}ccccccccc@{}}
\toprule
&   \multicolumn{4}{c}{Tough Tables}&\multicolumn{4}{c}{BiodivTab}  \\
\cmidrule(lr){2-5} \cmidrule(lr){5-9}
&   \multicolumn{2}{c}{CTA}& \multicolumn{2}{c}{CEA
}&\multicolumn{2}{c}{CTA} &\multicolumn{2}{c}{CEA} \\
\midrule
candidate number&   F1 & Pr & F1 &Pr &F1 & Pr & F1 & Pr \\
\cmidrule(lr){1-9}
1&   0.304& 0.333& 0.709&0.712&0.72& 0.73& 0.74& 0.76\\
 5&   0.537& 0.585& 0.826&0.827&0.86&0.86& 0.86&0.89\\
 10&   \textbf{0.596}& \textbf{0.629}& \textbf{0.843}&\textbf{0.845}&\textbf{0.89}& \textbf{0.89}& \textbf{0.90}&\textbf{0.93}\\
 15&0.587& 0.625& 0.830& 0.843& 0.88& 0.88& \textbf{0.90}&\textbf{0.93}\\
 
\toprule
\end{tabular}
\end{table}
\vspace{-20pt}

\subsubsection{Column Topic Detection}
% CEA列主题检测消融
Table \ref{tab:topic_comparison} shows the performance of the research system with and without column topic detection in CEA on the dataset. It can be seen that column topic detection has a positive effect on improving CEA. This may be because column topic detection can clearly define the core topic of each column of data, allowing the model to focus more precisely on relevant information, which helps the model better understand the data structure and semantic relationships. The model can quickly select and process valuable content based on the topic, thereby improving the precision.

% \vspace{-10pt}
\begin{table}[h]
\centering
\caption{Impact of column topic detection on System Performance}
\label{tab:topic_comparison}
\begin{tabular}{@{}ccccc@{}}
\toprule
&\multicolumn{2}{c}{Tough Tables}&\multicolumn{2}{c}{BiodivTab}\\
\cmidrule(lr){2-5} 
& \multicolumn{2}{c}{CEA
}&\multicolumn{2}{c}{CEA} \\
\midrule
System& F1 &Pr & F1 & Pr \\
\cmidrule(lr){1-5}
Our system with column topic detection& \textbf{0.843}&\textbf{0.845}& \textbf{0.90}&\textbf{ 0.93}\\
Our system without column topic detection& 0.815& 0.818& 0.89&0.90\\
 
\toprule
\end{tabular}
\end{table}
% \textbackslash{}vspace\{-5pt}

\subsubsection{Duplicate Removal}

Table \ref{tab:correct_comparison} shows the F1-scores and precision of CTA and CEA in Duplicate Removal. It can be seen that the effects of duplicate removal are slightly better. This may be because after duplicate removal, the data processed by the model is purer, and the system can focus on the truly valuable information and make more accurate judgments, thus improving the precision. After removing duplicate data, the amount of data decreases, and the computational load of the model is reduced, enabling the model to process data more efficiently. This allows the model to have more resources for more accurate analysis and judgment.
% 选不同k最好的结果
\vspace{-5pt}
\begin{table}[h]
\centering
\caption{Impact of Duplicate Removal on System Performance}
\label{tab:correct_comparison}
\begin{tabular}{@{}lcccc@{}}
\toprule
& \multicolumn{4}{c}{Tough Tables} \\
% \cmidrule(cr){2 - 5}
\cmidrule{2-5}
&   \multicolumn{2}{c}{CTA}& \multicolumn{2}{c}{CEA
} \\
\midrule
System&   F1 & Pr & F1 &Pr \\
\midrule
Our system with Duplicate Removal &\textbf{0.596}  &  \textbf{0.629} & \textbf{0.843}& \textbf{0.845}  \\
Our system without Duplicate Removal& 0.517& 0.544& 0.763& 0.768  \\
\bottomrule
\end{tabular}
\end{table}
\vspace{-20pt}

\subsubsection{CEA Based on \textit{Levenshtein Distance} }

To remove redundancy, we use to evaluate the performance of the CEA based on \textit{Levenshtein Distance} algorithm. We conducted tests on a subset of  the "Tough Tables" dataset by comparing the system's cell number with and without the algorithm. Table \ref{tab:algorithm1} shows that the system without the algorithm needs to process 177,355 cells, whereas the system with the algorithm only needs to process 60,341 cells—a difference of about 110,000 cells, representing a 2.83-fold increase. Our result clearly indicates that Algorithm 1 plays a crucial role in enhancing the system's data processing capability, significantly expanding its coverage and processing scale.

\vspace{-10pt}
\begin{table}[h]
\centering
\caption{Impact of Algorithm 1 on System Performance}
\label{tab:algorithm1}
\begin{tabular}{@{}lc@{}}
\toprule
\textbf{System} & \textbf{Cell Number} \\
\midrule
Our system with algorithm & 60,341\\
Our system without algorithm & 177,355\\
\bottomrule
\end{tabular}
\end{table}
\vspace{-20pt}

\section{Related works}

% 本研究使用LLMs ReAct 框架实现CTA和CEA任务，以完成OBDA中规则库的构建。下面将详细介绍同类型的其他方法，包括OBDA中（半）自动定义异构数据源之间的映射方法，CTA方法和CEA方法。

The following section provides a detailed overview of approaches about STA, including the CTA and CEA. STA involves five key tasks: CTA, CEA, Column-Property Annotation (CPA), Topic Annotation, and Row-to-Instance, with CTA and CEA being the primary focus. The SemTab challenge focuses on STA, with a particular emphasis on its application in linking tables to knowledge graphs\cite{semtab2020results,semtab2021results,semtab2022results,semtab2023results,semtab2024results}.

Existing STA methods span resource-efficient, structure-aware, and LLM-based designs. These approaches focus on candidate generation, accurate disambiguation, and leverage both table context and external knowledge to address the challenges of data complexity. To address the issue of hardware resource consumption, \cite{lowresource} proposed a low-resource system, which opted to create a custom data index. KGCODE-Tab\cite{kgcode} parsed the structure of the table, identifying the subject and non-subject columns. \cite{12} clustered candidate entities by leveraging geometric properties within a vector space. Dagobah\cite{dagobah} used a BERT-based hybrid model for entity disambiguation. s-elBat\cite{selbat} proposed an optimized search method to generate candidate entities. TorchicTab used RDF graph analysis and a pre-trained language model, LinkingPark utilized the modular specialized algorithms for candidate entity generation, entity disambiguation, property linking\cite{13,14}. Internal methods predicted intercell associations, while external methods inferred missing entities and relationships\cite{towardobda}. Kepler-aSI\cite{kepler} retrieved relevant entities and labels through SPARQL queries, matching entities using word embeddings and context information. A proposed anchoring model aligned data with ontology relationships, integrating symbolic reasoning, neural embeddings, and loss functions\cite{neuro}.

 Currently, there are some methods that use LLMs for CTA and CEA. These methods are aimed at semantic annotation of knowledge graphs and focus on the conversion from tables to knowledge graphs. \cite{ctawithgpt} was the first method used for CTA, achieving competitive results under few-shot conditions. ArcheType is an open-source framework that uses the LLMs for CTA, combining symbolic reasoning and neurals \cite{ctallm}.  
  \cite{table2kg} focused on matching tables containing only metadata to knowledge graph, using state-of-the-art methods in LLMs. CitySTI \cite{citysti} used LLMs to match tabular data with knowledge graphs, comparing the performance of Gemini, Llama, and GPT. Our method focuses on the STA from tables to ontology-based knowledge graphs. Moreover, it is implemented based on the ReAct-based agent, so it can be adapted to different table scenarios datasets. 

\section{Conclusion}

We propose a ReAct-based STA approach utilizing LLMs and external tools to effectively address CTA and CEA challenges. Experiments show superior performance on Tough Tables and BiodivTab datasets, achieving significant efficiency improvements with Levenshtein-based redundancy reduction (70\% less time, 60\% fewer tokens). Our method surpasses existing systems on the Tough Tables (CTA F1=0.596, CEA F1=0.843) and BiodivTab (CTA F1=0.89, CEA F1=0.90) datasets. The approach provides an automated, efficient, and cost-effective solution for semantic annotation of complex tables. The current method is applicable to general domains, future research will explore extending the approach to specialized domains.

% There will be a focus on further optimization in several aspects. Firstly, by enhancing semantic similarity methods and integrating domain-specific knowledge to better handle column topic annotation in the absence of contextual information. Secondly,our current method is applicable to general knowledge domains. The next step will be to explore semantic mapping for specialized domains. Thirdly, by adopting more efficient and cost-effective open-source large language models to enhance the scalability and practicality of the system. These improvements will make the system more robust and versatile in OBDA semantic mapping tasks.

\subsubsection*{Acknowledgment.}
This research project was supported in part by National Key Research and Development Program of China under Grant 2024YFB3312904, and Hubei Key Research and Development Program of China under Grant 2024BBB055, 2024BAA008; and in part by the Major Science and Technology Project of Yunnan Province under Grant 202502AE090003, and in part by the Fundamental Research Funds for the Chinese Central Universities under Grant 2662025XXPY005.

\bibliographystyle{splncs04} 

\bibliography{references}
% \appendix
% \input{appendix.tex}

\end{document}